%% file: emnlp2021.tex
\pdfoutput=1

\documentclass[11pt]{article}

\usepackage{acl}

\usepackage{times}
\usepackage{latexsym}

\usepackage[T1]{fontenc}

\usepackage[utf8]{inputenc}

\usepackage{microtype}

\usepackage{multirow}
\usepackage{array}
\usepackage{boldline}
\usepackage{bm}
\usepackage{filecontents}
\usepackage{amsfonts}
\usepackage{adjustbox}
\usepackage{amsmath}
\usepackage{mathtools}
\usepackage{color,soul}
\usepackage{esvect}
\usepackage{booktabs}
\usepackage{tikz}
\usepackage{graphicx}
\usepackage[fit]{truncate}
\DeclareGraphicsExtensions{.pdf,.png,.jpg}
\usepackage[normalem]{ulem}
\usepackage[ruled,vlined]{algorithm2e}
\usepackage{subfig}
\usepackage{float}
\usepackage{multicol}

\title{
Consistency Training with Virtual Adversarial Discrete Perturbation
}

\author{
Jungsoo Park$^{1,3} $\thanks{\, Equal contribution.} \quad  
Gyuwan Kim$^2$\footnotemark[1]\,\,\thanks{\, Work done while working at NAVER Clova \& AI Lab} \quad 
Jaewoo Kang$^3$ \\
$^1$Clova AI, NAVER Corp. \quad
$^2$ University of California, Santa Barbara \quad
$^3$Korea University \\
\texttt{\{jungsoo\_park,kangj\}@korea.ac.kr} \\
\texttt{gyuwankim@ucsb.edu}
}

\newcommand{\ours}{VAT-D}

\date{}

\begin{document}

\maketitle

\begin{abstract}
\input{00_abstract}

\end{abstract}

\input{01_introuction}
\input{02_method}
\input{03_experimental_setup}

\input{04_experimental_results}
\input{05_analysis}
\input{06_related_work}

\section*{Acknowledgements}
We thank Jinhyuk Lee, Jaewook Kang, and Sungdong Kim for the discussion and feedback on the paper. We also thank the members of the Conversation team in Naver CLOVA for active discussion. This research was supported by National Research Foundation of Korea (NRF-2020R1A2C3010638) and the Ministry of Science and ICT,  Korea, under the ICT Creative Consilience program (IITP-2022-2020-0-01819).

\bibliography{custom}
\bibliographystyle{acl_natbib}

\clearpage
\appendix
\numberwithin{table}{section}
\setcounter{page}{1}

\input{11_training_details_appendix}
\input{12_data_details_appendix}
\input{13_baseline_details_appendix}
\input{15_examples}



\end{document}

%% file: 00_abstract.tex
Consistency training regularizes a model by enforcing predictions of original and perturbed inputs to be similar.
Previous studies have proposed various augmentation methods for the perturbation but are limited in that they are agnostic to the training model.
Thus, the perturbed samples may not aid in regularization due to their ease of classification from the model.
In this context, we propose an augmentation method of adding a discrete noise that would incur the highest divergence between predictions. 
This virtual adversarial discrete noise obtained by replacing a small portion of tokens while keeping original semantics as much as possible efficiently pushes a training model's decision boundary. 
Experimental results show that our proposed method outperforms other consistency training baselines with text editing, paraphrasing, or a continuous noise on semi-supervised text classification tasks and a robustness benchmark\footnote{Code repo: \url{https://github.com/clovaai/vat-d}}.



%% file: 01_introuction.tex
\section{Introduction}

Building a natural language processing (NLP) system often requires an expensive process to collect a massive amount of labeled text data.
Semi-supervised learning (SSL)~\citep{chapelle2009semi} mitigates the requirement for such labeled data by exploiting the structure of unlabeled data.
Among the SSL methods, the consistency training framework~\citep{laine2016temporal,sajjadi2016regularization} enforces a model to produce similar predictions of original and perturbed inputs.
This method has several advantages over other training algorithms such as naively adding augmented samples into the training set~\citep{wei2019eda,ng2020ssmba} in that it provides a richer training signal than a one-hot label, and also applies to both labeled and unlabeled data~\citep{xie2019unsupervised}.

\begin{figure} [t!]
    \centering
     \subfloat[Real data distribution, which requires complex decision boundary.]
     {\includegraphics[width=0.45\columnwidth]{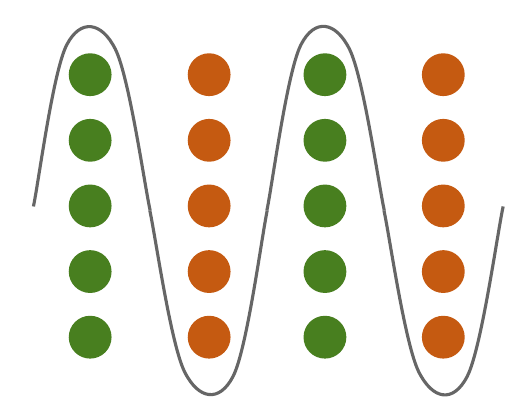}}\qquad
     \subfloat[A simple decision boundary is drawn when samples are insufficient.]
     {\includegraphics[width=0.45\columnwidth]{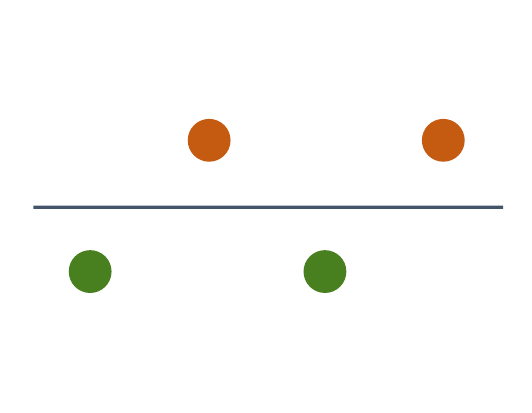}} \\
     \subfloat[Augmentations can push the decision boundary~(dotted line) from the current one~(bold line).]
     {\includegraphics[width=0.45\columnwidth]{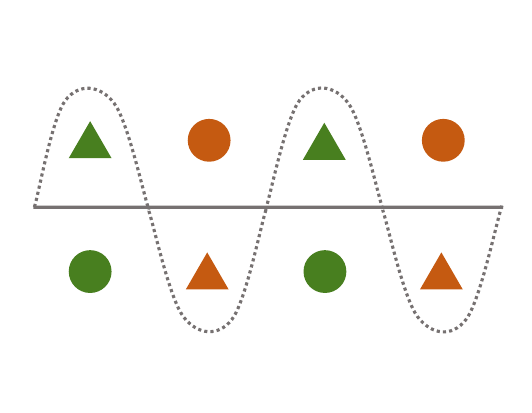}}\qquad
     \subfloat[Augmentations which are outside the current decision boundary enable further pushing it.]
     {\includegraphics[width=0.45\columnwidth]{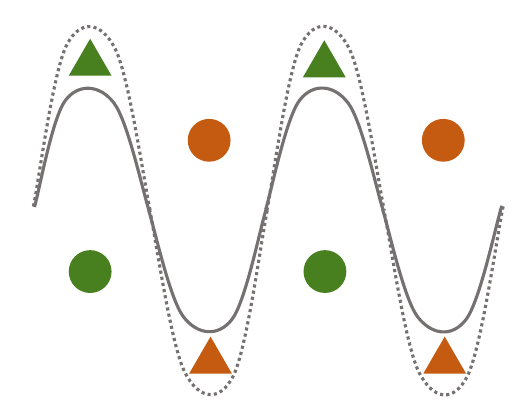}}
    \caption{
    A simple illustration of the intuition behind our method is visualized in the two-dimensional space, where the augmented samples~(triangle) would aid the training given a limited number of data~(circle).
    }
\label{fig:intro}
\end{figure}

For perturbing a text while preserving its semantics,
some approaches inject continuous noise to embedding vectors~\citep{xie2017data, miyato2018virtual}, and others modify text itself in discrete fashion by edit operations \cite{kobayashi2018contextual,wei2019eda} or paraphrasing with back-translation \cite{sennrich2015improving,edunov2018understanding, xie2019unsupervised}.  
However, adding continuous noise might not strongly regularize the training model, compared to diverse discrete noise-based augmentation methods~\citep{ebrahimi2017hotflip, cheng2019robust}.
Also, the augmentations with discrete noise are mostly black-box approaches based on simple rules or fixed models without access to the training model's internal states, having no control over output augmentations that would aid in the regularization of the training model.
As seen in Fig.~\ref{fig:intro}~(d), the augmented samples with similar semantics but that are outside the training model's decision boundary (\textit{i.e.} adversarial) are the ones that would effectively regularize the model to fit into the complex real data distribution.

To this end, we explore virtual adversarial training with discrete token replacements (\emph{\ours}). 
Our framework (1) first perturbs a given input text by replacing a small subset of tokens to \emph{maximize the divergence between the original and the perturbed samples' model predictions (i.e., virtual adversarial)} while filtering tokens to replace for constraining the semantic similarity, and (2) train a model to minimize the divergence of the predictions of original and perturbed inputs.

\ours{} shares the advantages of virtual adversarial training~(VAT) with continuous noise~\citep{miyato2018virtual} in that the perturbation is model-dependent, changing over the training time to approximate the augmented samples that would effectively push the decision boundary.
On the other hand, \ours{} differs from VAT in that the search space is discrete rather than continuous, thus not constrained by the pre-defined norm on the embedding space.
Our method relies on the training model's predictions which do not require label information, hence being the first work to successfully apply the adversarial training with perturbation on discrete space to the SSL framework. 

Our proposed method empirically outperforms previous state-of-the-art methods on topic classification datasets~\citep{chang2008importance, mendes2012dbpedia, zhang2015character} under various SSL scenarios. We additionally conduct experiments on ANLI robustness benchmark dataset~\citep{nie2019adversarial} for testing the robustness when only labeled samples are given where the method improves over the RoBERTa-Large~\citep{liu2019roberta} by 8 points.

%% file: 02_method.tex
\section{Background}

We explain the concept of consistency training and VAT that our framework relies on.

\paragraph{Consistency Training}

Consistency training~\citep{laine2016temporal, sajjadi2016regularization} enforces models' predictions to be invariant when the input is perturbed. 
This regularization pushes the decision boundary to traverse a low-density region~\citep{verma2019interpolation}.
The consistency loss is formally defined as
\begin{align}
\label{eq:consistency_regularization}
\mathcal{L}(\mathbf{x}, \mathbf{x'})=D[p(\cdot\mid\mathbf{x}), p(\cdot\mid\mathbf{x'})]
\end{align}

\noindent where $D$ is a non-negative divergence metric between two probability distributions~(\textit{e.g.}, KL-divergence), $\mathbf{x'}$ is a perturbed sample from an input $\mathbf{x}$ by any transformation.

\paragraph{Virtual Adversarial Training}

VAT~\citep{miyato2016adversarial, miyato2018virtual} is a consistency training method, which perturbs a given input with continuous noise to maximize the divergence from the model's prediction of the original input.
Such virtually adversarial examples effectively smooth the decision boundary compared to the random perturbation~\citep{miyato2018virtual}.
The formal definition of virtual adversarial samples is $\mathbf{\hat{x}} = \underset{\mathbf{x}' \in Neighbor(\mathbf{x})}{\operatorname{argmax}}{\mathcal{L}(\mathbf{x}, \mathbf{x'})}$
where the training objective is to minimize the $\mathcal{L}(\mathbf{x}, \mathbf{\hat{x}})$.
\citet{miyato2016adversarial} perturbs input by injecting noise to the embedding space, where the constraint of the perturbation is $\epsilon$-ball in $L^{p}$ norm centered at $\mathbf{x}$, \textit{i.e.} $Neighbor(\mathbf{x}) = \{\mathbf{x'} \mid \|\mathbf{x'}-\mathbf{x}\|_p \leq \epsilon\}$.


\section{Method}


We aim to generate a perturbed sample by adding discrete noise that incurs the highest divergence of the model's prediction logits from the original one without significant changes in its semantics. 
Our augmentation is made \textit{on-the-fly} depending on the current model to push the decision boundary during training effectively.

\paragraph{Virtual Adversarial Discrete Noise}
\label{sec:virtual_adversarial_word}


We develop the consistency training framework by perturbing inputs with virtual adversarial discrete noise, called \ours{}.
We want to perturb a given sentence $\mathbf{x} = (x_1, \dots, x_{M}) \in V^M$ of sequence length $M$ into a new sentence $\mathbf{x'} = (x'_1, \dots, x'_{M}) \in V^M$ of the same length, where $V$ is the word vocabulary.
In contrast with the continuous case, we constrain that $\mathbf{x'}$ differs from $\mathbf{x}$ in only small portion of positions changing their surface forms, \textit{i.e.} $Neighbor(x) = \{\|\mathbf{x'}-\mathbf{x}\|_H / M \leq \tau\}$ where $H$ denotes hamming distance in the token-level and $\tau$ is the replacement ratio.
In this work, we only focus on the replacement for simplicity.

\paragraph{Gradient Information}

The white-box approaches having an access to the training model's internal states, mostly rely on the gradient vectors of the loss function with respect to the input embeddings for finding adversarial discrete noise ~\citep{ebrahimi2017hotflip}.
However, for acquiring such gradient information under the framework of consistency training as in~Eq. \ref{eq:consistency_regularization}, naively resorting to the linear approximation of the loss function with respect to the input embeddings like in previous works~\citep{ebrahimi2017hotflip, michel2019evaluation, cheng2019robust} does not hold since the first-order term from Taylor expansion is zero when the label information is substituted to model's predictions~\citep{miyato2018virtual}. 

We bypass the obstacle by sharpening the distribution of original examples' predictions to enable the linear approximation.
Sharpening the distribution makes high probabilities higher and lower probabilities lower while not changing their relative order.
By sharpening the distribution of the original inputs' predictions, the first-order term does not result in zero, hence can be utilized for the approximation.
This is because the modified divergence loss is not zero when $\mathbf{x'}=\mathbf{x}$ indicating the non-negative divergence is not necessarily minimum at $r=\mathbf{x'}-\mathbf{x}=0$ (Note that the derivative of $f(x)$ is zero when the $f(x)$ is minimum at $x$).
The optimizing objective of Eq. \ref{eq:consistency_regularization} is modified to 
\begin{align}
\label{eq:modified_consistency_loss}
\mathcal{\Tilde{L}}(\mathbf{x}, \mathbf{x'}) = 
D[p^{\text{\textit{sharp}}}(\cdot\mid\mathbf{x})), p(\cdot\mid\mathbf{x}')] 
\end{align}
\noindent by sharpening the predicted distribution given an original input by the pre-defined temperature $T$ as $p^{\text{\textit{sharp}}}(\cdot\mid\mathbf{x}) = p(\cdot|\mathbf{x})^{\frac{1}{T}} / \left\|p(\cdot|~\mathbf{x})^{\frac{1}{T}}\right\|_{1}$.

\paragraph{Virtual Adversarial Token Replacement}

Consequently, the optimization problem to find a virtual adversarial discrete perturbation changes to
\begin{align*}
\mathbf{\hat{x}} = \underset{\mathbf{x}' \in Neighbor(\mathbf{x})}{\operatorname{argmax}}{\mathcal{\Tilde{L}}(\mathbf{x}, \mathbf{x'})}.
\end{align*}

\noindent Finally, we train the modified consistency loss function from Eq. \ref{eq:modified_consistency_loss} with obtained discrete perturbation.
The replacement operation of $m$-th token $x_m$ to the arbitrary token $x$ can be written as $\delta(x_m, x):=e(x)-e(x_m)$, where $e(\cdot)$ denotes embedding look-up.
We induce a virtual adversarial token by the following criteria~\citep{ebrahimi2017hotflip, michel2019evaluation, cheng2019robust, wallace2019universal, park2020adversarial}:
\begin{align}
\label{eq:vat_d_score}
\hat{x}_m =  \underset{x \in \operatorname{top\_k}(x_m, V)}{\operatorname{argmax}} \delta(x_m, x)^\top \cdot g_{x_m} \\
\label{eq:vat_d_gradient}
\text{where } g_{x_{m}} = \nabla_{e({x_{m})}}{\mathcal{\Tilde{L}}(\mathbf{x}, \mathbf{x'})}|_{\mathbf{x}'=\mathbf{x}} \nonumber
\end{align}
\noindent $\mathbf{g}_{x_m}$ is the gradient vector of the sharpened consistency loss from Eq. \ref{eq:modified_consistency_loss} with respect to the $m$-th token.
In brief, we replace the $m$-th original token $x_m$ with one of the candidates $x$ that approximately maximizes the consistency loss.
We randomly select token indexes to perturb and replace them simultaneously.
To bound the semantics similarity between the original sentence and the perturbed one, we use a masked language model~(MLM)~\citep{devlin2019bert, liu2019roberta} to restrict a set of possible candidates to replace $x_m$.
We filter top-k candidates~\citep{cheng2019robust}, denoted as $\operatorname{top\_k}(x_m, V)$, from the vocabulary having the highest MLM probability at position $m$ when an original sentence $x$ is given to the MLM. 
More training details are in Appendix~\ref{app:training_details}.

%% file: 03_experimental_setup.tex
\section{Experimental Setup}

\input{tables/ssl_results}
\input{tables/anli_results}

\subsection{Dataset}

We experiment on three topic classification datasets and Adversarial NLI (ANLI)~\citep{nie2019adversarial}.
The former evaluate our method's effectiveness in SSL and the latter is for evaluating the robustness of the models under the standard supervised training framework.
The three topic classification benchmarks consist of AG News
~\citep{zhang2015character}, DBpedia~\citep{mendes2012dbpedia}, and YAHOO! Answers~\citep{chang2008importance}.
We follow the experimental setting from~\citet{chen2020mixtext}, where we train with a limited number of labeled data in diverse settings, namely, 10, 200, 2500 per class.
We randomly sample the labeled, unlabeled, and development set and report the performance on the official test set.
For producing the confident results, we report the average of five different seeds' distinct runs.

As for ANLI, we train the model with two different settings, training with only the ANLI dataset or additionally training with other NLI datasets, including SNLI~\citep{bowman2015large}, MNLI~\citep{williams2017broad}, and FEVER~\citep{thorne2018fever} following the original work~\citep{nie2019adversarial}. Further details are in Appendix~\ref{app:data_details}.

\subsection{Baseline}

We compare our method with various baselines of the perturbation methods including EDA~\citep{wei2019eda}, UDA~\citep{xie2019unsupervised}, VAT~\citep{miyato2016adversarial, miyato2018virtual} for the topic classification SSL task.
For the ANLI dataset, we compare with the baselines~\citep{devlin2019bert,  yang2019xlnet,  liu2019roberta, jiang2019smart} that have reported numbers on the official validation and test set. 
More details are in Appendix~\ref{app:baselines}.

\subsection{Training Details}

We exploit the unlabeled data from the topic classification datasets and the labeled data from the ANLI for consistency loss. 
Throughout the experiments, we set the replacement ratio $\tau$ as 0.25 and top-k as 10.
We sharpen the predictions with $T$ as 0.5 for topic classification datasets~(including baselines) and 0.75 for the ANLI.

%% file: tables/ssl_results.tex
\begin{table*}[t!]
\centering
\resizebox{0.75\textwidth}{!}{
\def\arraystretch{0.7}
\begin{tabular}{ l | ccc | ccc | ccc }
\toprule
\multicolumn{1}{l|}{\multirow{2}{*}{\textbf{Method}}} & \multicolumn{3}{c|}{\textbf{AG\_NEWS}} & \multicolumn{3}{c|}{\textbf{YAHOO!}} & \multicolumn{3}{c}{\textbf{DBpedia}} \\[1pt] 
& \textit{\textbf{10}}   & \textit{\textbf{200}}  & \textit{\textbf{2500}} & \textit{\textbf{10}}   & \textit{\textbf{200}}  & \textit{\textbf{2500}} & \textit{\textbf{10}}   & \textit{\textbf{200}}   & \textit{\textbf{2500}}   \\ [1pt]\midrule
 BERT~\citep{devlin2019bert}                    & 79.4 & 88.6 & 91.6 & 58.2 & 70.1 & 73.9 & 97.8 & 98.8  & 99.1    \\ [3pt] 
 EDA~\citep{wei2019eda}                         & 83.8 & 88.9 & 91.8 & 62.0 & 70.6 & 73.8 & 98.4 & 98.8  & 99.1    \\ [3pt] 
 UDA~\citep{xie2019unsupervised} & 83.8 & 88.5  & 91.6 & 62.0 & 70.4 & 73.7 & 98.2 & 98.8  & 99.1                  \\ [3pt] 
  VAT~\citep{miyato2016adversarial}             & 82.3 & 88.9 & 91.8 & 62.4 & 70.7 & 74.1 & 98.4 & 98.8  & 99.1    \\ [1pt] 
\midrule
 VAT-D                                          & \textbf{86.2} & \textbf{90.0} & \textbf{92.3} & \textbf{65.3} & \textbf{71.7} & 74.1 & 98.4 & \textbf{99.0} & \textbf{99.2}           \\ 
\bottomrule
\end{tabular}}
\caption{Accuracy on topic classification datasets under the various SSL settings. 
10, 200, 2500 denote the number of labeled samples per class used during training. 
We average five different runs with a differently indexed dataset to show the significance~\citep{dror2018hitchhiker}.
The numbers in the bold denote the best score.
}
\label{tab:ssl_results}
\vspace{-.15cm}
\end{table*} 

%% file: tables/anli_results.tex
\begin{table*}[t!]
\centering
\resizebox{0.70\textwidth}{!}{
\def\arraystretch{0.85}
\begin{tabular}{ l | cccc | cccc}
\toprule
\multicolumn{1}{l|}{\multirow{2}{*}{\textbf{Method}}} & \multicolumn{4}{c|}{\textbf{Dev}} & \multicolumn{4}{c}{\textbf{Test}} \\[3pt]
& A1 & A2 & A3 & ALL & A1 & A2 & A3 & ALL \\[1pt] \midrule
\multicolumn{9}{c}{\textbf{MNLI + SNLI + ANLI + FEVER}} \\ [1pt]\midrule
BERT\citep{nie2019adversarial}      & 57.4 & 48.3 & 43.5 & 49.3 & - & - & - & 44.2  \\
XLnet~\citep{nie2019adversarial}    & 67.6 & 50.7 & 48.3 & 55.1 & - & - & - & 52.0  \\
RoBERTa~\citep{nie2019adversarial}  & 73.8 & 48.9 & 44.4 & 53.7 & - & - & - & 49.7  \\
SMART~\citep{jiang2019smart}        & \textbf{74.5} & 50.9 & 47.6 & 57.1 & \textbf{72.4}  & 49.8 & \textbf{50.3} & 57.1  \\
VAT-D                               & \textbf{74.5} & \textbf{54.2} & \textbf{50.8} & \textbf{59.2} & \textbf{72.4}  & \textbf{51.8} & 49.5 & \textbf{57.4}  \\ [1pt] \midrule
\multicolumn{9}{c}{\textbf{ANLI}} \\ [1pt] \midrule
RoBERTa~\citep{nie2019adversarial}  & 71.3 & 43.3 & 43.0 & 51.9 & - & - & - & -  \\
SMART~\citep{jiang2019smart}        & 74.2 & 49.5 & 49.2 & 57.1 & \textbf{72.4} & 50.3 & 49.5 & 56.9  \\ 
VAT-D                               & \textbf{74.8} & \textbf{52.1} & \textbf{51.1} & \textbf{58.8} & 72.1 & \textbf{51.4} & \textbf{51.7} & \textbf{57.9}   \\
\bottomrule
\end{tabular}}
\caption{Accuracy on the ANLI benchmark. 
The numbers of the baselines are from the original papers~\citep{nie2019adversarial, jiang2019smart}.
The upper section is for training with all the NLI datasets, and the bottom is for training with only the ANLI.}
\label{tab:anli_results}
\vspace{-.25cm}
\end{table*}

%% file: 04_experimental_results.tex
\section{Experimental Results}

\subsection{Semi Supervised Text Classification}
\label{sec:result_ssl}

Table \ref{tab:ssl_results} shows the experimental results on topic classification datasets under SSL setup.
Our method outperforms the baselines by up to 7.1 points from the BERT model finetuned with standard cross-entropy loss and 2.9 points from other methods utilizing the consistency regularization loss.
The accuracy gained from the proposed method from the baselines, especially when the number of labeled samples is limited. 
However, since all the methods have already achieved high accuracy in the DBpedia, the difference among methods is not significant.

Among the baselines, VAT~\citep{miyato2016adversarial,miyato2018virtual} performs reasonably well.
The finding supports the claim that a transformation during consistency training should be done with regard to the training model.

\subsection{Adversarial Natural Language Inference}

Table \ref{tab:anli_results} shows the experimental results on the ANLI dataset with different training settings: training with all the NLI datasets, or training with only the ANLI dataset.
Our method improves over baselines, including RoBERTa-Large~\citep{liu2019roberta} and SMART~\citep{jiang2019smart} in both settings. 
Specifically, our method improves on an average of 8.0 points in the test set from training with cross-entropy loss only.
Compared to SMART, which combines smoothness regularization, i.e., a variation of VAT, and Bregman proximal point optimization for finetuning, our method outperforms it on an average of 1 point from the test set without using other techniques such as Bregman proximal point optimization.

%% file: 05_analysis.tex
\section{Effectiveness of the White-box Search}
\label{sec:adv_search_analysis}

\input{tables/search}

\begin{figure}[!t]
    \centering
     \subfloat[Top-k Distribution]
     {\includegraphics[width=0.5\columnwidth]{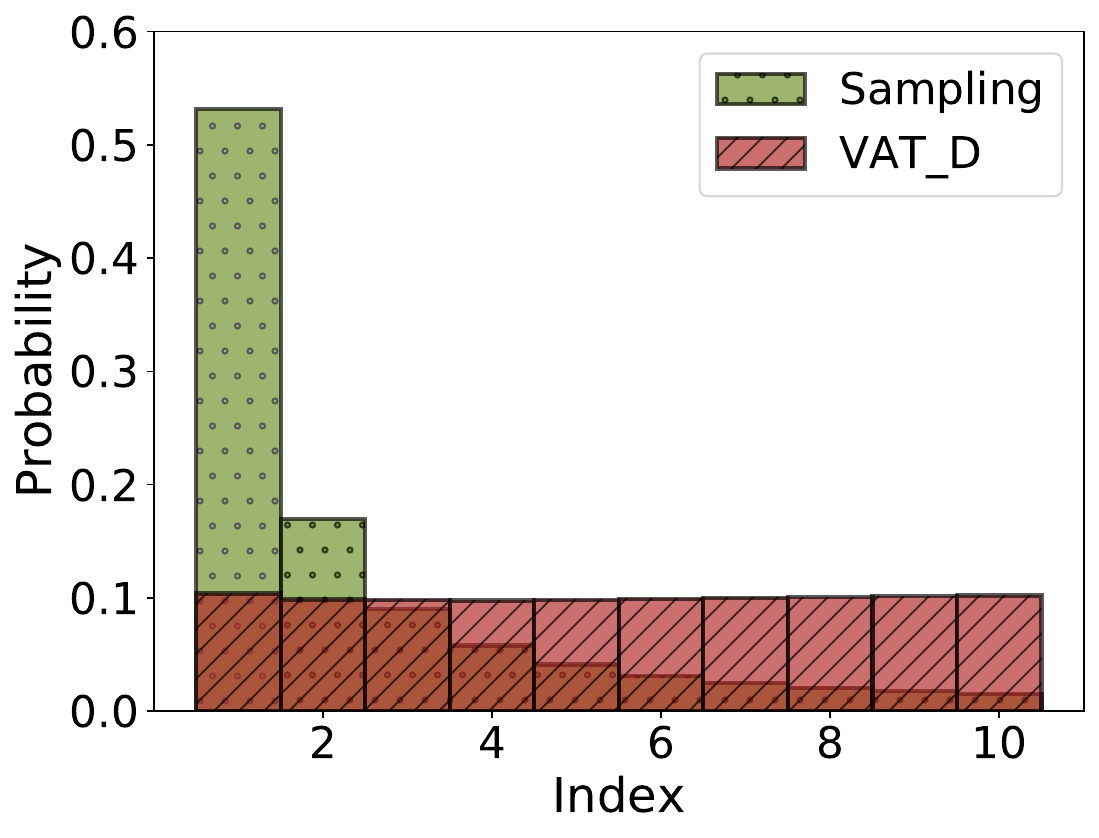}}
     \subfloat[Consistency Loss]
     {\includegraphics[width=0.5\columnwidth]{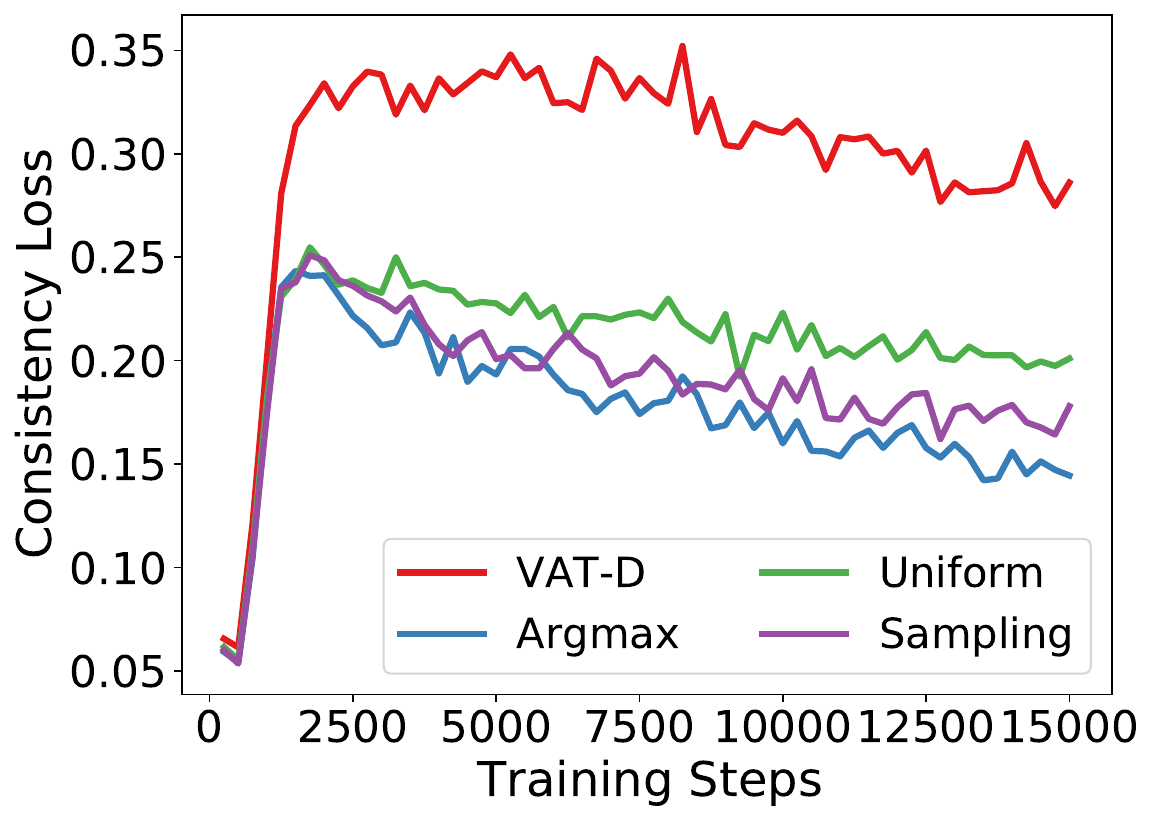}}
    \caption{Index distribution from the top-k candidates sorted by MLM scores (a) and consistency loss of different sampling strategies during training (b).}
    \label{fig:density_loss}
    \vspace{-.25cm}
\end{figure}

Our central intuition behind the proposed method is to generate the augmented samples concerning the model, i.e., vulnerable to the model.
This section further conducts an ablation study on whether such virtual adversarial search is crucial in discrete space.
We select the token among the top-K candidates that would incur the highest divergence from the model's prediction.
We compare with other sampling strategies among the top-K candidates, namely, uniform sampling (Uniform), selecting the token with maximum MLM probability (Argmax), and sampling from MLM probabilities (Sampling).
We match other training details except for the sampling strategy for a fair comparison.

Table~\ref{tab:search} illustrates the result of comparisons.
The virtual adversarial search among candidates outperforms other search strategies in discrete space, especially when the number of labeled samples is limited. The result demonstrates that the virtual adversarial search is indeed the crucial component during perturbation.
Furthermore, Fig.~\ref{fig:density_loss} shows the indexes that our method selected from top-k distribution~(sampling from YAHOO! dataset) during training.
The distribution of indexes selected from our method resembles the uniform distribution; however, as in the loss plot from Fig.~\ref{fig:density_loss}, our method searches for the diverse yet adversarial candidates to the model, i.e., incurring high divergence.

%% file: tables/search.tex
\begin{table}[!t]
\centering
\resizebox{\columnwidth}{!}{
\def\arraystretch{0.85}
\begin{tabular}{ l | ccc | ccc}
\toprule
\multicolumn{1}{l}{\multirow{2}{*}{\textbf{Method}}} & \multicolumn{3}{c}{\textbf{AG\_NEWS}} & \multicolumn{3}{c}{\textbf{YAHOO!}} \\ 
& \textit{\textbf{10}}   & \textit{\textbf{200}}  & \textit{\textbf{2500}} & \textit{\textbf{10}}   & \textit{\textbf{200}}  & \textit{\textbf{2500}}   \\ [1pt] \midrule
VAT-D    & \textbf{86.2} & \textbf{89.8} & \textbf{92.3} & \textbf{65.3} & \textbf{71.7} & \textbf{74.1} \\ [1pt] \midrule
Uniform  & 83.8 & 89.3 & 91.8 & 63.2 & 70.8 & 73.8  \\ 
Argmax   & 83.2 & 89.0 & 91.9 & 63.7 & 70.9 & 73.7  \\
Sampling & 84.8 & 89.3 & 91.8 & 63.5 & 70.9 & 73.8  \\
\bottomrule
\end{tabular}}
\caption{
\label{tab:search}
Accuracy according to different sampling strategies from top-K candidates.
}
\vspace{-.3cm}
\end{table}

%% file: 06_related_work.tex
\section{Related Works}

\paragraph{Consistency Regularization} 
Consistency regularization~\citep{laine2016temporal, sajjadi2016regularization} has been mainly explored in the context of SSL~\citep{chapelle2009semi, oliver2018realistic}.
A line of research in text-domain~\citep{miyato2016adversarial, clark2018semi, xie2019unsupervised, miyato2018virtual, jiang2019smart, asai2020logic} explored the idea.
Existing studies explored varying perturbation methods.
Injecting norm-constrained continuous noise to the embedding space~\citep{miyato2016adversarial, jiang2019smart, liu2020adversarial, chen2020seqvat, sano2019effective} and directly perturbing the text~\citep{clark2018semi, minervini2018adversarially, li2019logic, xie2019unsupervised, asai2020logic} via discrete noise are the primary approaches for the perturbation. 
Our method perturbs the sentence by the discrete noise, yet the noise is generated concerning the training model.

\paragraph{Adversarial Training} 
Our method extends the white-box-based adversarial training framework~\citep{Goodfellow2014adv, madry2018towards}, which has recently been explored widely in NLP~\citep{miyato2016adversarial, ebrahimi2017hotflip, michel2019evaluation, wang2019robust, zhu2019freelb, jiang2019smart, liu2020adversarial}.
\citet{cheng2019robust} use adversarial training on machine translation by discrete word replacements relying on the label information, so not applicable to SSL different from ours.
There are also black-box approaches for generating the adversarial attacks or test sets~\citep{jia2017adversarial, alzantot2018generating, ribeiro2018semantically, ribeiro2019red, gardner2020evaluating} to evaluate the vulnerability of the NLP models, unlike our method, which utilizes gradient information during training.
\citet{li2020bert, garg2020bae, li2021contextualized} perturb input using MLMs similar to ours but designed for an attack so inefficient for adversarial training.

\paragraph{Data Augmentation} 
Synthetically generated training examples are utilized to augment an existing dataset~\citep{feng2021survey}. 
Existing word-level augmentation methods~\citep{zhang2015character, xie2017data, wei2019eda} are based on heuristics. 
Mixup-based methods~\citep{zhang2018mixup} interpolate input texts in hidden embeddings~\citep{chen2020mixtext, guo2019augmenting} or input-level \cite{yoon2021ssmix, kim2021linda}.
Other methods include utilizing back-translation models~\citep{sennrich2015improving, xie2019unsupervised}, contextual language models~\citep{kobayashi2018contextual, wu2019conditional},
or generative models~\citep{anaby2020not, yang2020g}.
Unlike previous works, our method is subject to the training model, thus approximating the augmented points, efficiently filling in gaps from the training data.

%% file: 11_training_details_appendix.tex
\section{Training Details}
\label{app:training_details}

\input{algorithms/vawr}

Alg.~\ref{alg:VAdv_WR} illustrates the procedure (VAT\_D) to acquire virtual adversarial tokens with the modified consistency loss.
We randomly select token indexes to perturb $I$, subject to the length of the sentence.
Considering multiple substitutions, an exhaustive search over all possible combinations to find the optimal one is computationally intractable.
For efficient generation during each training step, we replace multiple tokens simultaneously instead of greedy search or beam search, which has shown to work considerably well in previous works~\citep{ebrahimi2017hotflip, cheng2019robust}.
During training, the models are optimized with standard cross-entropy and consistency loss with an equal weight where we utilize KL-Divergence as the divergence $D$. 
Our method takes approximately 2.5 times the standard training whereas other baselines (e.g., EDA, Back-translation) take about 1.7 times the standard training.
We utilize P40 for training the SSL experiments and V100 for the ANLI task.

In our preliminary experiment, utilizing the MLM with masking was worse than that without masking, similar to \citet{li2020bert}.
While utilizing the MLM for filtering top-k candidates, we empirically verified that not applying masking operations to the sentence achieved better performance than doing so.
We conjecture that the loss of information when applying masking operation has evoked the perturbed samples to significantly deviate from the original ones, resulting in a degradation in performance.
The finding matches that of \citet{li2020bert}.
Thus we do not apply masking operations throughout the experiments.
Moreover, we do not fine-tune the \textit{off-the-shelf} MLM on the training corpus but only the classification model, which is to ensure a fair comparison with other augmentation baselines.

%% file: algorithms/vawr.tex
\setlength{\textfloatsep}{10pt}

\begin{algorithm}[t!]
\SetKwInOut{Input}{Input}
\SetKwInOut{Output}{Output}
\SetKwProg{Fn}{Function}{:}{}
\Input{$\text{input sentence}~\mathbf{x}, \text{index to perturb}~\mathbf{I} $ 
}

\Output{$\text{perturbed sentence}~\mathbf{\hat{x}}$}
\Fn{VAT\_D({$\mathbf{x}$, $\mathbf{I}$})}{
$\mathbf{\hat{x}} \gets \mathbf{x}$\\ 
\For{$ m \in \mathbf{I}$}{
    $g_{x_{m}} \gets \nabla_{e({x_{m})}}{\mathcal{\Tilde{L}}(\mathbf{x}, \mathbf{x'})}|_{\mathbf{x}'=\mathbf{x}}$ \\
    $\hat{x}_m \gets \underset{x \in \text{top\_k}(x_m, V)}{\operatorname{argmax}} \delta(x_m, x)^\top \cdot ~g_{x_m}$ \\
    Replace $m$-th token of $\mathbf{\hat{x}}$ to $\hat{x}_m$
}
}
\Return {$\mathbf{\hat{x}}$}
\caption{VAT\_D Module}
\label{alg:VAdv_WR}
\end{algorithm}

%% file: 12_data_details_appendix.tex
\section{Further Details on Data}
\label{app:data_details}

\input{tables/data_statistics_ssl}
\input{tables/data_statistics_nli}

The dataset statistics and split information regarding topic classification tasks and ANLI is presented in Table~\ref{tab:data_statistics_ssl} and Table~\ref{tab:data_statistics_nli}.

ANLI~\citep{nie2019adversarial} is an NLI testbed recently introduced for evaluating the robustness of the models in natural language understanding.
The dataset consists of three rounds (A1-A3), each consisting of a train-dev-test set with increasing difficulty, where the data is generated by human-and-model-in-the-loop fashion to fool the strong pre-trained models~\citep{devlin2019bert,  yang2019xlnet, liu2019roberta}.

%% file: tables/data_statistics_ssl.tex
\begin{table}[t!]
\centering
\resizebox{0.42\textwidth}{!}{
\begin{tabular}{ lccccc}
\toprule
\textbf{Dataset}  & \textbf{Genre} & \textbf{Class} & \textbf{Unlabel} & \textbf{Dev} & \textbf{Test}   \\ \midrule
AG\_NEWS & News        &  4   & 20k & 20k & 19k       \\ 
YAHOO!   & QA          &  10  & 50k & 20k & 60k       \\ 
DBPedia  & Wikipedia   &  14  & 70k & 20k & 50k       \\ \bottomrule 
\end{tabular}}
\caption{Data statistics for the topic classification datasets following the experimental setting from~\citet{chen2020mixtext}.}
\label{tab:data_statistics_ssl}
\end{table} 

%% file: tables/data_statistics_nli.tex
\begin{table}[t!]
\centering
\resizebox{0.42\textwidth}{!}{
\begin{tabular}{ l  cccc  }
\toprule
\textbf{Dataset}  & \textbf{Genre} &  \textbf{Train} & \textbf{Dev} & \textbf{Test}   \\ \midrule
A1    & Wikipedia       &   17k  & 1k & 1k          \\ 
A2    & Wikipedia       &   45k  & 1k & 1k          \\ 
A3    & Various         &   100k & 1.2k & 1.2k     \\ 
ANLI  & Various         &   162k & 3.2k & 3.2k     \\ \midrule
MNLI  & Various         &   392k & - & -              \\ 
Fever & Wikipedia       &   208k & - & -              \\ 
SNLI  & Image Captions  &   549k & - & -              \\ \bottomrule 
\end{tabular}}
\caption{Data statistics for the ANLI with three rounds~(A1-A3) and concerning NLI datasets for the training.}
\label{tab:data_statistics_nli}
\end{table}

%% file: 13_baseline_details_appendix.tex
\section{Further Details on Baselines}
\label{app:baselines}

For the SSL setup, we use the following baselines:

\paragraph{BERT~\citep{devlin2019bert}}
We use the pre-trained BERT-base-uncased model and finetune it for the classification dataset using only standard cross-entropy loss. 

\paragraph{EDA~\citep{wei2019eda}}
EDA is a simple data augmentation strategy based on word unit operations such as synonym replacement or deletion.
We perturb the unlabeled samples using EDA\footnote{\url{https://github.com/jasonwei20/eda_nlp}} and exploit them for consistency training.

\paragraph{UDA~\citep{xie2019unsupervised}} 
UDA paraphrases the sentence using the back-translation. 
We employ the WMT-19 DE$ \leftrightarrow $EN model from fairseq\footnote{\url{https://github.com/pytorch/fairseq}}~\citep{ott2019fairseq} to do the back-translation on unlabeled samples, and exploit them for consistency training.

\paragraph{VAT~\citep{miyato2016adversarial, miyato2018virtual}} 
We re-implement VAT where we apply the consistency loss to the unlabeled samples.

%% file: 15_examples.tex
\section{Augmentation Quality}

\begin{table*}[t!]
\centering
\resizebox{\textwidth}{!}{
\def\arraystretch{0.85}
\begin{tabular}{ l | l}
\toprule
Source & Sample \\ \midrule
\multicolumn{2}{l}{AG\_NEWS} \\ \midrule
Original & Turkey agonized over pressure to recognize cyprus in the \hl{last} hurdle to an historic \hl{agreement} \\
Augmentation & Turkey agonized over pressure to recognize cyprus in the \hl{final} hurdle of an historic \hl{deal} \\
Original & Rockets \hl{struck} a baghdad hotel housing \hl{foreign} contractors and journalists late thursday \\
Augmentation & Rockets \hl{hit} a baghdad hotel housing \hl{visiting} contractors and journalists late thursday \\
Original & Ten \hl{people} were injured yesterday when a bomb exploded outside the \hl{Indonesian} embassy in Paris \\
Augmentation & Ten \hl{civilians} were injured yesterday when a bomb exploded outside the \hl{Jakarta} embassy in Paris \\
Original & Pakistan \hl{authorities} are putting the city of Karachi \hl{on} ... for an al - qaida \hl{strike} after its \hl{forces} killed \hl{a} top \hl{terror} suspect \\
Augmentation & Pakistan \hl{governments} are putting the city of karachi \hl{of} ... for an al - qaida \hl{bomb} after its \hl{members} killed \hl{one} top \hl{terrorism} suspect  \\

\midrule
\multicolumn{2}{l}{YAHOO!} \\ \midrule
Original & How can \hl{guests} get sound security under wireless internet environment at hotel \hl{?} \\
Augmentation & How can \hl{visitors} get sound security under wireless internet environment at hotel \hl{...} \\
Original & \hl{Can} you find \hl{some} ones screen name by using there real \hl{name} ? ? ? yes \\
Augmentation & \hl{Could} you find \hl{other} ones screen name by using there real \hl{surname} ? : ? yes \\
Original & What \hl{is} the perfect gift for my girlfriends b - day ? ? ? ( information in here about her ) ? she \hl{loves} to : ride your black \hl{sport} bike \\
Augmentation & What \hl{will} the perfect gift for my girlfriends b - day ? ? ? ( information here here of her ) ? she \hl{wants} to : ride your black \hl{racing} bike
\\
Original & Purpose of administration and it department to a \hl{business} ? i work in \hl{it} ... without us , companies would be at a \hl{standstill} \\
Augmentation & Purpose of administration and it department to a \hl{corporation} ? i work in \hl{that} ... without us of companies would be near a \hl{stands market} \\

\midrule
\multicolumn{2}{l}{DBpedia} \\ \midrule

Original & Patryk \hl{Dominik} Sztyber ( \hl{born} \hl{4} august 1979 in Opoczno ) stage name seth is a \hl{Polish} heavy metal musician \\
Augmentation & Patryk \hl{Deinik} Sztybor ( \hl{birth} \hl{8} august 1979 in Opoczno ) stage name seth is a \hl{Warsaw} heavy metal musician \\
Original & Twill is a quarterly magazine \hl{published} between Paris and Milan. It has an international \hl{readership} \\
Augmentation & Twill is the quarterly magazine \hl{printed} between Paris and Milan. It has an international \hl{readers range} \\
Original & The \hl{pond} \hl{creek} station \hl{located} \hl{east} of Wallace Kansas ... is a two - story frame building that was a stagecoach station \hl{built} 1865 \\
Augmentation & The \hl{lake} \hl{branch} station \hl{built} \hl{outside} to Wallace Kansas ...is a two - story frame building that been a stagecoach station \hl{designed} 1865 \\
Original & Until we have wings is an album by randy stonehill \hl{released} in \hl{1990} on myrrh records \\
Augmenatation & Until we have wings is an album by randy stonehill \hl{published} mid \hl{1989} on myrrh records \\

\bottomrule
\end{tabular}}
\caption{Generated augmentation examples from our method along with original samples}
\label{tab:aug_results}
\end{table*}

We present some augmentation samples in Table~\ref{tab:aug_results} from three topic-classification datasets. As presented in the table, the augmentation samples moderately modify some tokens from the original sentence following the original context. 

However, since we are decoding multiple tokens at a same time, some samples are shown to be ungrammatical (e.g., \emph{is} $\rightarrow$ \emph{will} instead of \emph{will be}). Moreover, if the chosen token to be modified are entities, the augmentation sample can sometimes change the information presented in the sentence (e.g., \emph{Patryk Dominik} $\rightarrow$ \emph{Patryk Deinik}). However, since we are solving the task of the closed-domain topic classification task, the problems didn't matter much in this setting.
If we are to solve the knowledge-intensive task, we would have to consider other filtering modules for not changing the entities.